\documentclass[10pt,journal,compsoc]{IEEEtran}
\usepackage[english]{babel}
\usepackage{ragged2e}

\usepackage{epsfig}
\usepackage{graphicx}
\usepackage{subfigure}
\usepackage{multirow}
\usepackage{color}
\usepackage{algorithm}
\usepackage{algorithmic}
\usepackage{color}

\ifCLASSOPTIONcompsoc
  \usepackage[nocompress]{cite}
\else
  \usepackage{cite}
\fi

\ifCLASSINFOpdf

\else

\fi
\begin{document}

\title{AI Oriented Large-Scale Video Management for Smart City: Technologies, Standards and Beyond}
\author{Lingyu Duan$^{1}$, Yihang Lou$^{1,2}$, Shiqi Wang$^{3}$,Wen~Gao$^{1,2}$,~\IEEEmembership{Fellow,~IEEE}, Yong Rui$^{4}$,~\IEEEmembership{Fellow,~IEEE} 

{\small\begin{minipage}{\linewidth}\begin{center}
\begin{tabular}{ccc}
$^{1}$Institute of Digital Media, Peking University, Beijing, China \\
$^{2}$SECE of Shenzhen Graduate School, Peking University, Shenzhen, China \\
$^{3}$Department of Computer Science, City University of Hong Kong \\
$^{4}$Lenovo Research, Beijing, China \\
\end{tabular}
\end{center}\end{minipage}}
}


%
\maketitle
%
%

\begin{abstract}
Deep learning has achieved substantial success in a series of tasks in computer vision. Intelligent video analysis, which can be broadly applied to video surveillance in various smart city applications, can also be driven by such powerful deep learning engines. To practically facilitate deep neural network models in the large-scale video analysis, there are still unprecedented challenges for the large-scale video data management. Deep feature coding, instead of video coding, provides a practical solution for handling the large-scale video surveillance data. To enable interoperability in the context of deep feature coding, standardization is urgent and important. However, due to the explosion of deep learning algorithms and the particularity of feature coding, there are numerous remaining problems in the standardization process. This paper envisions the future deep feature coding standard for the AI oriented large-scale video management, and discusses existing techniques, standards and possible solutions for these open problems.
\end{abstract}


\section{Introduction}
\label{sec:intro}
Recently, a considerable number of deep learning algorithms have been proposed, which exhibit substantial performance improvement in various computer vision tasks.
Compared with traditional handcrafted features, deep learning algorithms aim to learn representative features from the vast amounts
of training data. Since AlexNet \cite{witten2016data} won the ImageNet competition, there are tremendous research activities focusing on designing more powerful and deeper networks. Follow ups like VGGNet \cite{simonyan2014very} , GoogleNet  \cite{szegedy2015going}, ResNet \cite{he2016deep} and DenseNet \cite{huang2016densely} have greatly improved the discrimination capability of features to a higher level, which also boosted the performance of many visual analysis tasks. Generally speaking, these technologies have naturally made substantial impact on public security, such as face recognition \cite{liu2017sphereface}, person \cite{xiao2016learning} and vehicle reidentification \cite{liu2016deep} in surveillance videos.

Recent years have witnessed dramatically increased demand for the smart city construction, where the concerning safety issues have received sufficient interest.
In particular, there is a vast and increasing proliferation of surveillance videos acquired and transmitted over both wireline and wireless networks.
Due to the real-time recording of the physical world, surveillance video is very valuable and there is considerable concern regarding
how to efficiently manage such surveillance video big data. In view of the explosion of the surveillance systems deployed in urban areas and millions of objects/events captured every day,
there are a unique set of
challenges regarding efficient analysis and
search. In particular, video compression and transmission constitute the basic infrastructure to support these applications.
Though the state-of-the-art video coding standards such as H.265/HEVC have dramatically
improved the coding performance, it is still questionable that whether such big video data can be efficiently handled by visual signal level compression.
Fortunately, an alternative strategy ''analyze then compress'' provides a solution, which transmits the compact features extracted and compressed at the edge end to the server side.
Such paradigm can sufficiently satisfy various intelligent video analysis tasks, by using significantly less data than the compressed video itself. In Fig. 1, the infrastructure of the smart city with large-scale video management based on feature extraction and transmission is illustrated. In particular, to meet the demand of large-scale video analysis in smart city applications, the feature stream instead of video signal stream can be transmitted. As such, the intelligent front end devices extract features locally and then convey the encoded feature stream to the server for analysis purpose. 

While the field of artificial intelligence is still quickly evolving, and efficient and novel
deep learning algorithms will continue to emerge in the coming years, it is also interesting to discuss how we could enable the interoperability
of the compressed deep learning features in real-world applications. In contrast with video coding, which directly compresses the visual signals into the bitstream, feature coding involves both feature extraction and compression process.
In particular, feature extraction serves as the raw features producer to generate the source for compression and is responsible for the answer of what to compression. Feature compression accounts for the conversion of raw deep features into compact representation bitstream.
The purpose of this article is to provide an overview of the existing deep learning techniques in video surveillance and envision the future deep learning feature coding standards. We will start by a
brief review of the current status of deep learning in video surveillance, followed by discussions on the compact feature standard in MPEG. Then the open problems of deep feature coding standardization will be discussed, where we can perceive both great
promises and challenges.

\begin{figure*}[]
\centering
\includegraphics[width=7.3in]{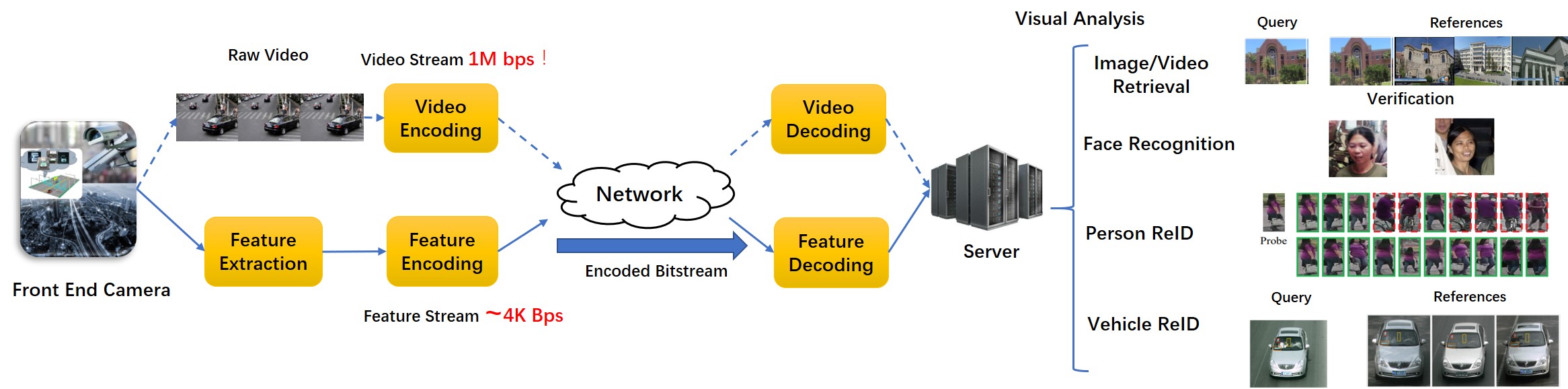}
\caption{The infrastructure of the large-scale video management with feature transmission for smart city applications. 
}
\label{fig:cdva_pipeline}
\end{figure*}
\section{Standardization of Compact Handcrafted Feature Descriptor}

\subsection{Compact Descriptors for Visual Search}
In view of the importance of the transmission of feature descriptors, MPEG has finalized the standardization of Compact Descriptors for Visual Search (CDVS) \cite{duan2016overview} and published the standard in Sep. 2015. In CDVS, handcrafted local and global descriptors are leveraged to represent the visual characteristics of images. The normative blocks of CDVS involve the extraction of local and global descriptors. More specifically, the local descriptors consist of SIFT descriptors which are efficiently compressed by a low-complexity transform coding. The raw local descriptors are further selected and aggregated to generate a Scalable Compressed Fisher Vector (SCFV), with competitive matching accuracy and low memory footprint. In view of the fluctuation of available bandwidth in the mobile environment, CDVS supports interoperability between different size bitstream by setting six operating points from 512B to 16KB. More technical details about CDVS are referred to \cite{duan2016overview}.

\subsection{Compact Descriptors for Video Analysis}
The emerging requirements of video analysis facilitate the standardization of large-scale video analysis. The MPEG has moved forward to standardize Compact Descriptors for Video Analysis (CDVA) \cite{mpeg_cdva_cfp}. Video consists of sequentially correlated frames, such that extracting the feature from each frame leads to high redundancy in feature representation and unnecessary computational costs.  
The ongoing CDVA standard adopted multi-keyframe based image retrieval, which converts the problem of video retrieval into an image retrieval task. In particular, the local and global descriptors of standardized CDVS descriptors are extracted on the sampled keyframes of a given query video, which are further packed together to constitute the CDVA descriptors. Moreover, the deep learning features \cite{lin2017hnip} are also adopted to further boost the analysis performance.  Fig.~\ref{fig:cdva_pipeline} presents the framework of ongoing CDVA with handcrafted features and deep features, including the video structure and the normative components of feature extraction. It is also worth mentioning that the NIP descriptor has been adopted into the working draft of CDVA standard. 

\section{Deep Learning in Video Surveillance}
\label{sec:review}
With the exponential growth of the video surveillance data, video content analysis has been a long standing research topic in computer vision community. Currently, there are four urgent visual analysis tasks in the surveillance scenario, i.e., image/video retrieval, person Re-Identification, face recognition and vehicle retrieval. These tasks play important roles in building the safety city and ensuring the public security. 
Associated with these tasks, the core techniques in establishing the visual system of the smart city are shown in Table 1. Multiple academic disciplines, including visual signal processing, computer vision, compression as well as hardware architectures are involved in the further construction of the system. It is envisioned that with the standadlization of deep learning features and the advancements of these technologies, the system that sees intelligently, efficiently and greenly will eventually come true.

\begin{table}[]
\centering
\caption{The core techniques involved in the visual system of smart city.}
\label{my-label}
\begin{tabular}{|l|l|}
\hline
The Core Techniques                                                  & Description                                                                                                                                                 \\ \hline
\begin{tabular}[c]{@{}l@{}}Feature \\ Generation\end{tabular}        & Extract discriminative feature representation                                                                                                               \\ \hline
\begin{tabular}[c]{@{}l@{}}Feature\\ Generalization\end{tabular}     & \begin{tabular}[c]{@{}l@{}}Generalize the deep feature, to different\\ tasks e.g., from person ReID to vehicle ReID.\end{tabular}                           \\ \hline
\begin{tabular}[c]{@{}l@{}}Feature Redudancy \\ Removal\end{tabular} & \begin{tabular}[c]{@{}l@{}}Remove the redudancies of the features in \\ spatial or temporal domain\end{tabular}                                             \\ \hline
\begin{tabular}[c]{@{}l@{}}Rate-Distortion \\ Optimization\end{tabular} & \begin{tabular}[c]{@{}l@{}}Investigate the distortion of features \\ and optimize feature compression with\\  rate-distortion optimization.\end{tabular} \\ \hline
\begin{tabular}[c]{@{}l@{}}Feature \\ Binarization\end{tabular}      & \begin{tabular}[c]{@{}l@{}}Binarize features to enable fast feature \\ transmission and analysis.\end{tabular}                                              \\ \hline
\begin{tabular}[c]{@{}l@{}}Network \\ Compression\end{tabular}       & \begin{tabular}[c]{@{}l@{}}Efficiently represent the network and \\ lower the disk and memory cost.\end{tabular}                                            \\ \hline
\end{tabular}
\end{table}
\subsection{Image/Video retrieval} 
Image/video retrieval refers to searching for the images/videos representing the same objects or scenes as the one depicted in the query, which may present under different scales, illuminations, rotations or even occlusions. In the last decade, the image/video retrieval has benefited a lot from handcrafted SIFT descriptors due to its robustness to the image transformations. However, after the AlexNet \cite{witten2016data} won ILSVRC12 by a significant margin, the CNN-based image feature representation has become mainstream techniques when handing complex and semantic vision analysis. 
Regarding to both image and video retrieval, competitive and even better retrieval performance  \cite{babenko2015aggregating} \cite{babenko2015aggregating}  \cite{tolias2015particular} \cite{lin2017hnip} has been reported on several benchmarks. 

CNN-based retrieval methods can be categorized into two types: pre-trained and fine-tuned CNN models. 
The commonly used pre-trained CNN models are trained on ImageNet dataset consisting of 1.2 million images of 1000 classes, such that the features can be regarded as generic. The descriptors can be extracted from fully-connected (FC) layers or intermediate layers. The FC descriptors have a global receptive field and the intermediate local descriptors have a smaller receptive field and location information encoded in 2D feature maps. To obtain the global representation, encodings like VLAD and FV are usually adopted. In addition, the direct pooling can also generate discriminative features. For example, in \cite{tolias2015particular} Tolias et al. employed max pooling on selected regions in intermediate feature maps and subsequently performed sum pooling. Though impressive results can be achieved by pre-trained model, there is a trend to fine-tune CNN model on a task-oriented dataset for specific retrieval. The classification and verification based networks are two typical types. The former is trained to classify pre-defined categories, and later adopts siamese network \cite {gordo2016deep} with contrastive loss or triplet loss. On several retrieval benchmarks such as Holidays, Oxford5K, Paris 6K, the fine-tuned models have achieved the state-of-the-art performance. 

\begin{figure}[htbp]
\centering
\includegraphics[width=0.95\linewidth]{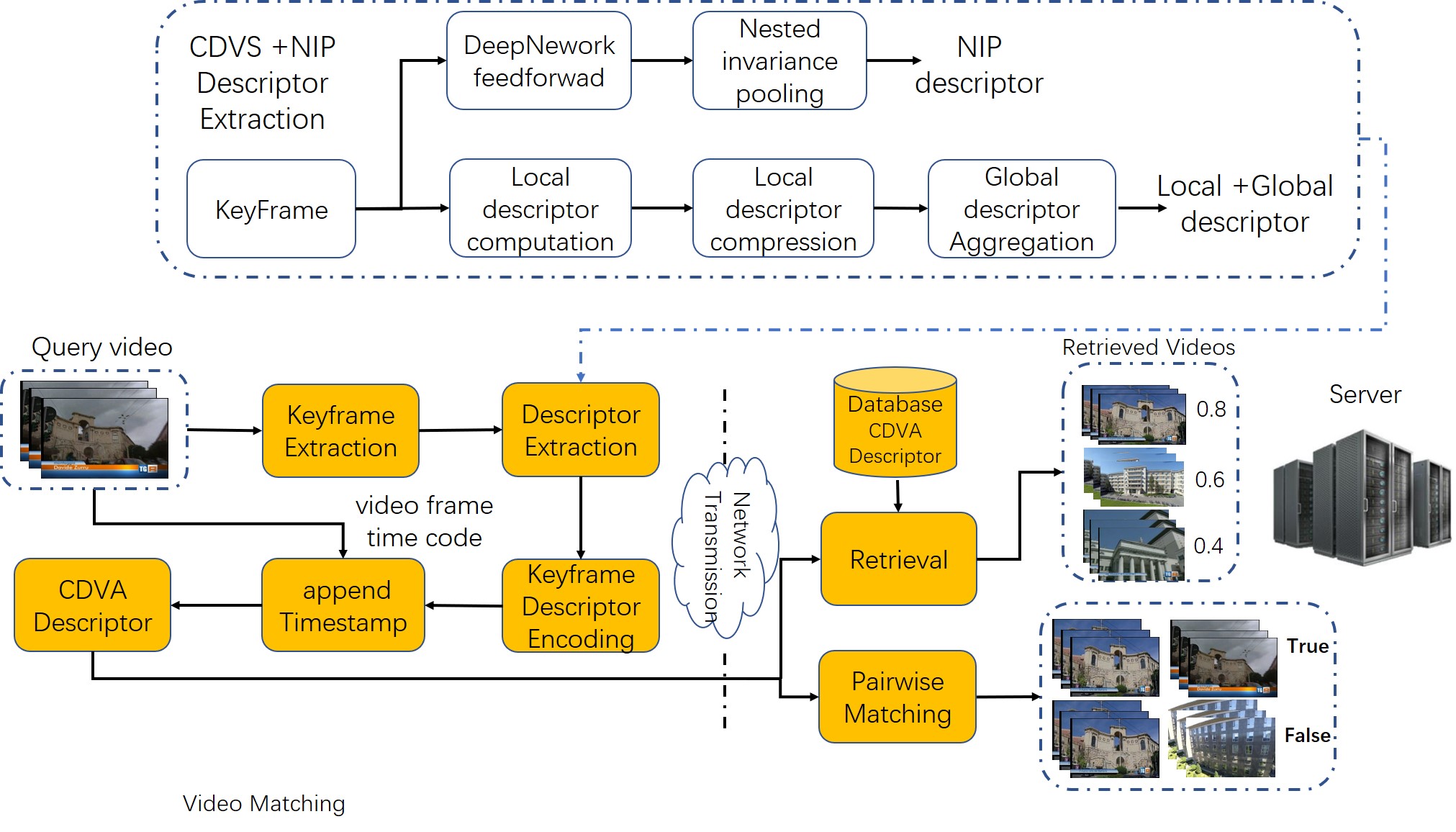}
\caption{ Illustration of MPEG CDVA evaluation framework}
\label{fig:cdva_pipeline}
\end{figure}

\subsection{Person ReIdentification}
Person ReIdentification (ReID) has attracted more and more research focus due to its application significance in video surveillance. It aims to search whether a given person is present in other cameras. The widespread camera deployment in public places and the increased safety requirements make the existing manual labor spotting scheme powerless when facing the real-time generated massive video data. A practical person ReID system involves person detection, tracking, and retrieval. In particular, person retrieval is the main research focus among the related works. The challenge of this task is how to accurately match two images of the same person under variant scenes, viewpoints, scales and lighting conditions.

Deep ReID systems mostly employ two types of CNN models, i.e., siamese and classification models. The difference between these two models lies in the input form and definition of the loss function. The siamese models \cite{Su2016Deep} \cite{Chen2016Similarity} leverage image pairs or triplets as input, then let them forward propagate to get feature vector in embedding space. The distance of images with the same person is constrained on the feature vector by a minimum margin using contrastive loss or triplet loss. By contrast, the classification models \cite{Matsukawa2016Hierarchical} \cite{zhao2017spindle} treat each person identity as a class, and the classification based loss functions are usually employed, such as softmax loss. These models pay more attention to the feature representation from the perspective of local and global combination, part-attention model, human body's skeleton model, etc. Intuitively, the combinations of siamese and classification model have also received a lot of attention.


\subsection{Face recognition}
Due to the nonintrusive recognition manner (intrusive like fingerprint, retina recognition), face recognition has great application potential in surveillance security. 
Over the last decades, amount of works in face recognition \cite{liu2017sphereface} \cite{wen2016discriminative} have emerged, which greatly boosted the accuracy on the popular benchmark such as Labeled Face in the Wild (LFW) to an unprecedent level. In real-word applications, the captured face images may not be as high quality as that in LFW dataset, creating many challenging problems originated from arbitrary poses, low quality resolutions, occlusions and small scales.

Typically, face recognition includes face identification and face verification. The former classifies a given face as a specific identity, and the latter verifies whether a given face pair belongs to the same identity. Regarding experimental setup, there are closed-set and open-set settings. Under closed-set, the testing identities are contained in the training set. By contrast, in open-set, the testing identities do not appear in the training set. Therefore, the real-world recognition can be regarded as face verification in open-set. Essentially, this task is defined as a metric learning problem. The expected feature representation should be able to meet the demand of small intra-class distance and large inter-class distance. Deep models are capable of building the above criterion by setting appropriate loss functions. 

\subsection{Vehicle ReIdentification}

In many vehicle-relevant tasks, vehicle ReIdentification is the most crucial technique in city security. The license plate is usually the straightforward choice to identify a vehicle. However, in real applications, most surveillance cameras are not equipped with recognition capability. Furthermore, the license plates of vehicles in many cases are occluded or faked. Thus, the visual appearance based techniques present great application prospect. Compared with the classic person Re-identification problem, vehicle ReIdentification is more challenging since it faces the enormous inter-class similarity and intra-class variances presented by massive vehicles of the same model types and the shooting situation variations across multiple cameras. For example, the subtle differences between similar vehicles are even challengeable for human beings. Fortunately, some special marks such as tissue box, pendant, annual inspection marks, etc provide characteristics clues for efficient discrimination. 

The deep metric learning has been widely adopted for Vehicle ReID tasks. The objective of the deep network is to learn a deep embedding where the samples of the same vehicle ID are constrained in a local space, such that the samples of different vehicle ID are farther away than ones of the same vehicle ID. Such feature distribution 
is pretty desirable for nearest neighbor retrieval. As such, the retrieval method is the main solution for re-identification. As mentioned above, the granularity of vehicle ReID is finer than person ReID. Consequently, the inter-class feature distribution requires more structured prior knowledge to represent such subtle differences. This motivated some recent attempts which incorporate intra-class variance into the feature representation, such as group sensitive triplet embedding in \cite{bai2017incorporating}.





\begin{figure*}[htbp]
\label{retrieval1}
\begin{minipage}[t]{0.23\linewidth}
\includegraphics[width=1\linewidth]{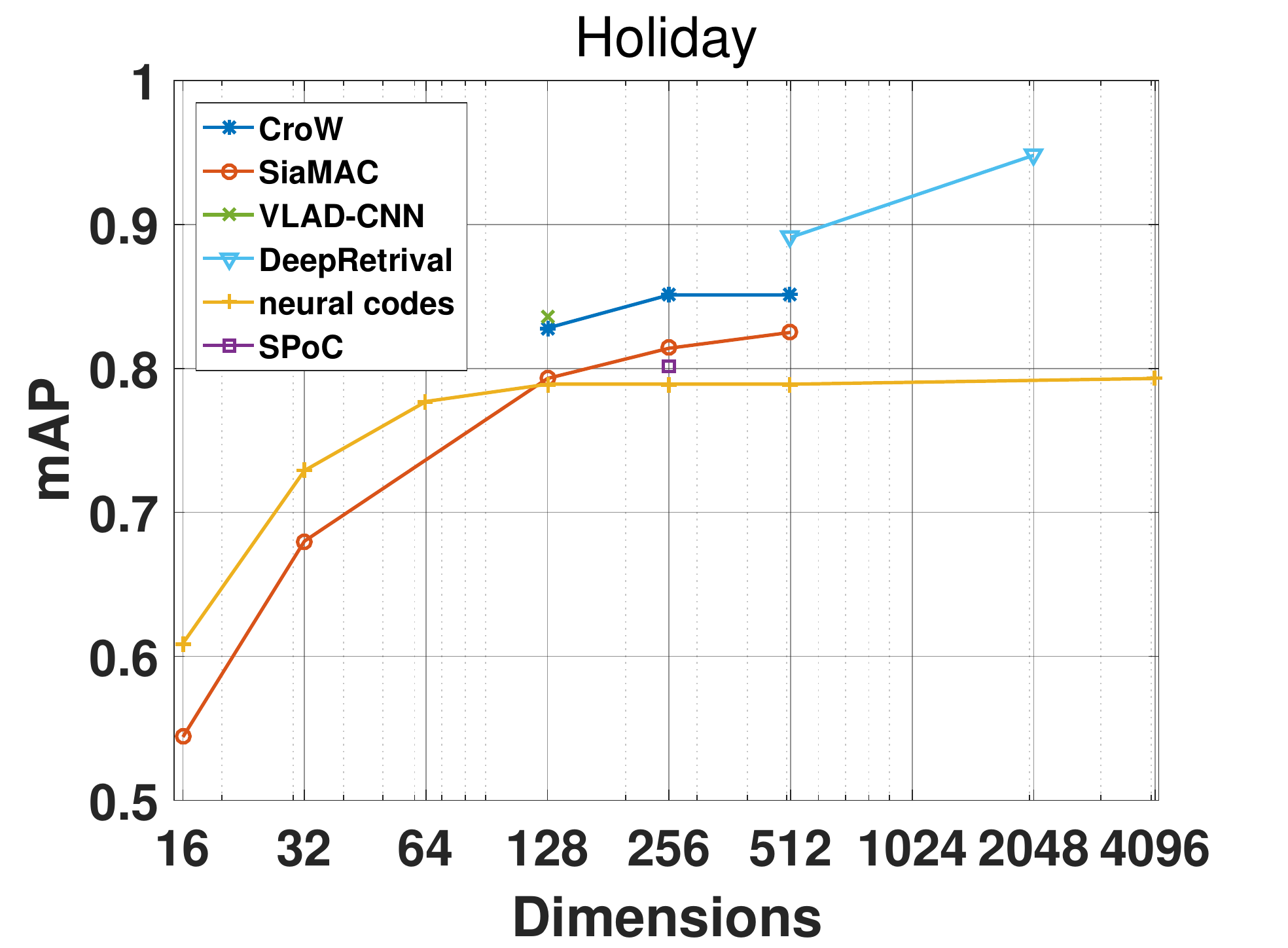}
 \centerline{(a)}  
\end{minipage}
\hfill
\begin{minipage}[t]{0.23\linewidth}
\includegraphics[width=1\linewidth]{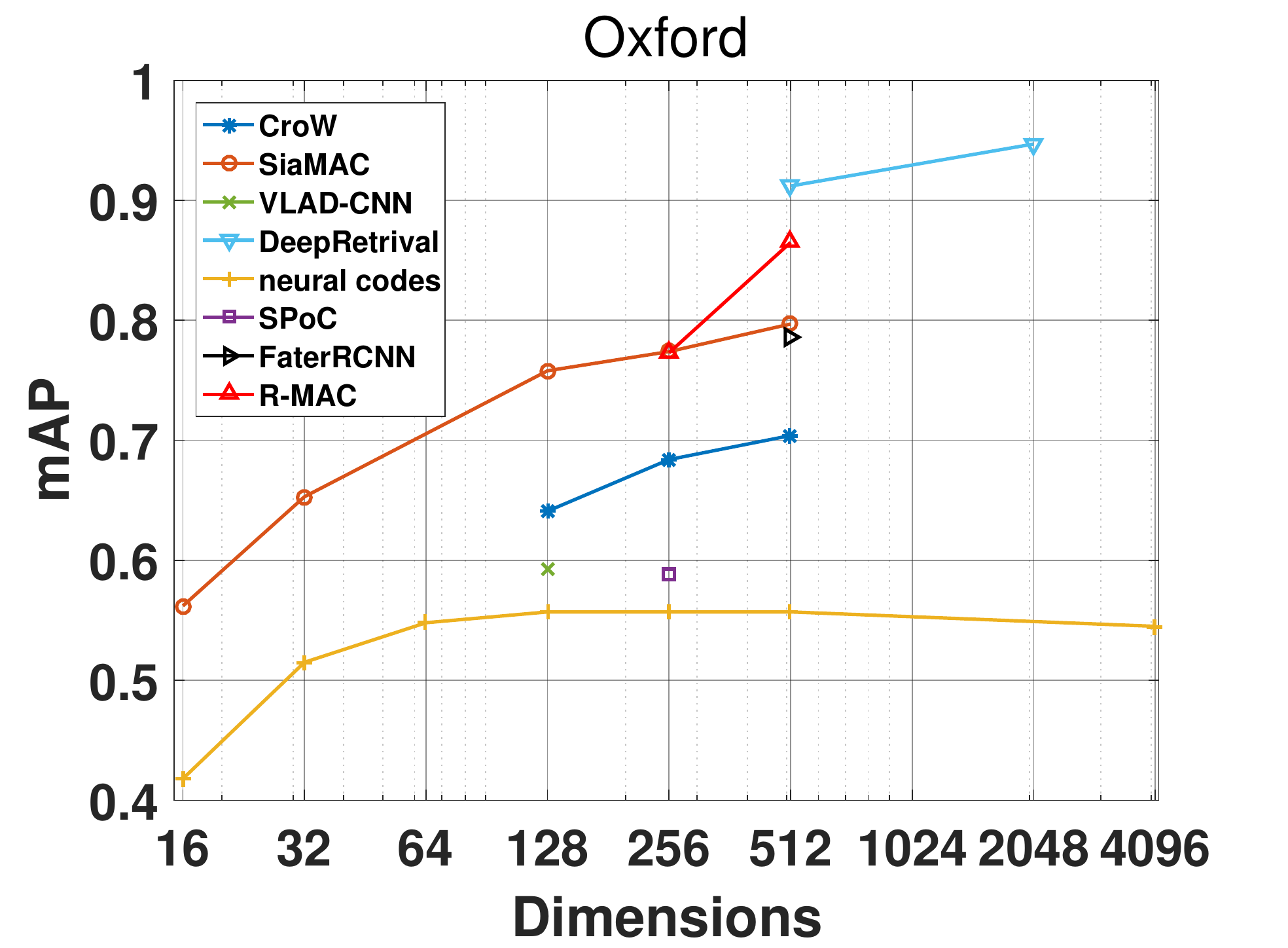}
\centerline{(b)}  
\end{minipage}
\hfill
\begin{minipage}[t]{0.23\linewidth}
\includegraphics[width=1\linewidth]{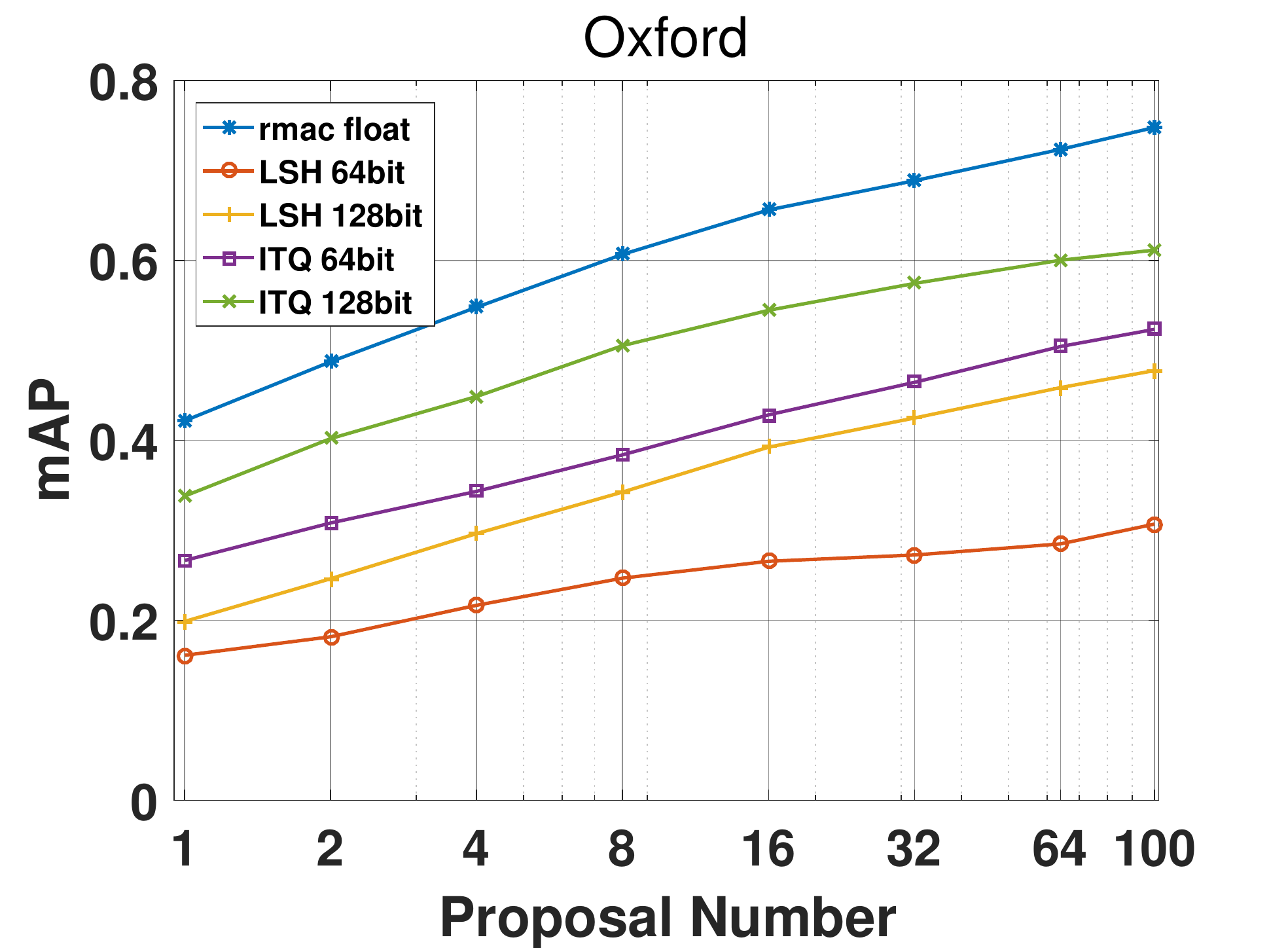}
\centerline{(c)}  
\end{minipage}
\hfill
\begin{minipage}[t]{0.23\linewidth}
\includegraphics[width=1\linewidth]{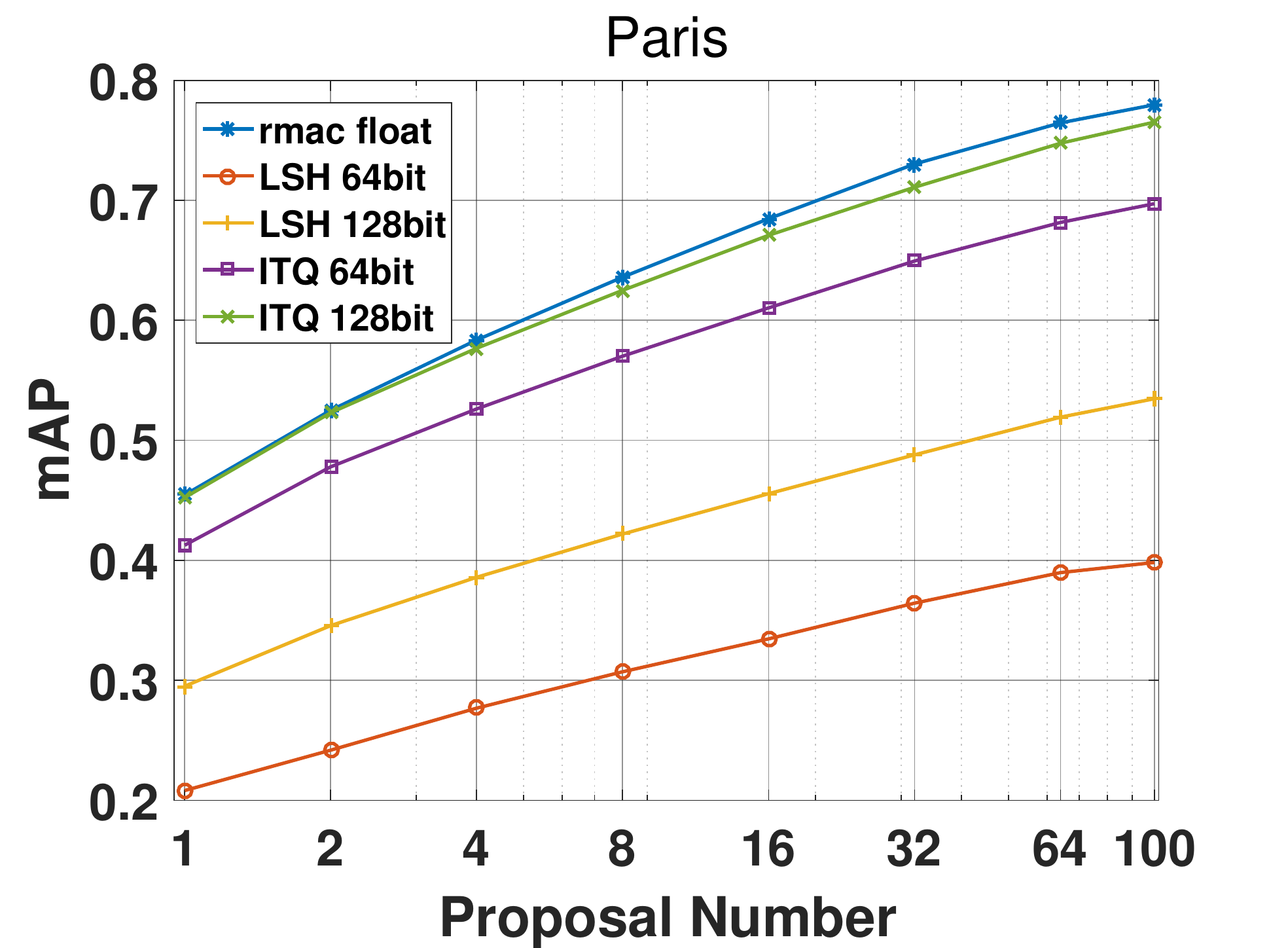}
\centerline{(d)}  
\end{minipage}
\caption{Performance variations with the increase of dimensions and object proposals for recent works on image retreival benchmarks.}
\end{figure*}

\begin{table}[]
\centering
\caption{Performance comparisons of methods reported on CDVA benchmarks where the landmarks, scene and objects are the sub datasets}
\label{cdva}
\begin{tabular}{|c|c|c|c|c|c|c|}
\hline
Methods                & Dims & Year & \begin{tabular}[c]{@{}l@{}}Land\\ mark\end{tabular} & Scene & Objects & All   \\ \hline
CXM0.2   \cite{mpeg_cdva_16_05_cxm}   & 1024 & 2016 & 0.598     & 0.594 & 0.917   & 0.721 \\ \hline
MAC    \cite{tolias2015particular}            & 512  & 2015 & 0.619     & 0.762 & 0.718   & 0.670 \\ \hline
SPoC    \cite{babenko2015aggregating} & 256  & 2015 & 0.691     & 0.840 & 0.703   & 0.709 \\ \hline
CroW   \cite{babenko2015aggregating}  & 512  & 2015 & 0.639     & 0.784 & 0.720   & 0.683 \\ \hline
R-MAC   \cite{tolias2015particular}  & 512  & 2015 & 0.746     & 0.873 & 0.782   & 0.771 \\ \hline
HNIP VGG \cite{lin2017hnip}    & 512  & 2016 & 0.748     & 0.901 & 0.850   & 0.801 \\ \hline
HNIP Alex \cite{lin2017hnip} & 768  & 2016 &   - &    - &     - & 0.772 \\ \hline
HNIP Res \cite{lin2017hnip} & 2048 & 2016 &   -   &   - &    - & 0.817 \\ \hline
\end{tabular}
\end{table}

\begin{table}[]
\centering
\caption{Performance comparisons of vehicle re-identification methods on VehicleID and VeRI benchmarks}
\label{vehicleid}
\begin{tabular}{|c|c|c|c|c|}
\hline
Methods                   & Dims & Year & VEHICLEID & VERI-776 \\ \hline
Triplet Loss   \cite{Wang2014Learning}     & 400  & 2015 & 0.373     &    -      \\ \hline Softmax Loss       & 1024 & 2015 & 0.580     & 0.343    \\ \hline
Triplet+Softmax Loss \cite{bai2017incorporating}  & 1024 & 2014 & 0.650     & 0.558    \\ \hline
BOW-CN                & 100  & 2015 &    -       & 0.122    \\ \hline
CCL VGGM    \cite{liu2016deep}           & 1024 & 2016 & 0.386     &   -       \\ \hline
Mixed Diff+CCL  \cite{liu2016deep}   & 1024 & 2016 & 0.455     &     -     \\ \hline
HDC+Contrastive  \cite{Yuan2016Hard}     & 384  & 2017 & 0.575     &    -      \\ \hline
GSTE  \cite{bai2017incorporating}   & 1024 & 2017 & 0.724     & 0.594    \\ \hline
\end{tabular}
\end{table}

\begin{table}[]
\centering
\caption{Performance summarization of some representative Person ReID methods on the benchmarks}
\label{person}
\begin{tabular}{|c|c|c|c|c|c|c|}
\hline
Methods    & Dims & Year & \multicolumn{1}{c|}{VIPeR} & \multicolumn{1}{c|}{CUHK1} & \multicolumn{1}{c|}{PRID} & \multicolumn{1}{c|}{GRID} \\ \hline
Cov-of-Cov \cite{Serra2014Covariance} & 16828      & 2014 & 33.9                       & 40.9                        & 47                            & 16.6                      \\ \hline
LOMO \cite{Liao2015Person} & 26960      & 2015 & 40                               &  -                     & 15.3                          & 16.6                      \\ \hline
GOLD  \cite{Serra2015GOLD}     & 1169       & 2015 & 27.1                       & 35.3                        & 40.5                          & 10.9                      \\ \hline
2AvgP \cite{Carreira2015Free}  & 952        & 2015 & 28.8                       & 36.1                        & 44.7                          & 12.9                      \\ \hline
GOG-RGB  \cite{Matsukawa2016Hierarchical}  & 7567       & 2016 & 42.3                       & 55.8                        & 63.6                          & 22.8                      \\ \hline
NFST  \cite{Zhang2016Learning}     & 5138       & 2016 & 51.2                       & 69                          &  -                            & -                          \\ \hline
SCSP  \cite{Chen2016Similarity}     & 120        & 2016 & 53.5                       & 24.2                        &  -                             & -                          \\ \hline
SSDAL  \cite{Su2016Deep}    & 105        & 2016 & 43.5                       &  -                           & 20.1                          & 19.1                      \\ \hline
TMA    \cite{Martinel2016Temporal}    & 100        & 2016 & 39.9                       &  -                           & 54.2                          & -                          \\ \hline
P2S     \cite{zhou2017point}   & 800        & 2017 &            -                & 77.3                        &  -                            & -                          \\ \hline
Spindle  \cite{zhao2017spindle}  & 256        & 2017 & 53.8                       & 79.9                        & 67                            & -                          \\ \hline
\end{tabular}
\end{table}

\begin{table}[]
\centering
\caption{A summary of efficiency and accuracy comparisons between recent remarkable works on face recognition}
\label{face}
\begin{tabular}{|c|c|c|c|c|}
\hline
Methods & Year & Dims & LWF   & YTF   \\ \hline
DeepFace \cite{taigman2014deepface} & 2014          & 4096       & 97.35 & 91.4  \\ \hline
Learning Face \cite{sun2014deep} & 2014          & 10575      & 97.73 & 92.2  \\ \hline
MDML-DCPs \cite{parkhi2015deep}  & 2015          & 1024       & 98.95 & 97.3  \\ \hline
FaceNet     \cite{schroff2015facenet}   & 2015          & 128        & 99.63 & 9512  \\ \hline
Deep embedding \cite{liu2015targeting}      & 2015       & 128        & 99.13 &    -   \\ \hline
Multimodal deep face \cite{ding2015robust} & 2015       & 9000       & 98.43 &    -   \\ \hline
Center Loss  \cite{wen2016discriminative}    & 2016        & 512        & 99.28 & 94.9  \\ \hline
Large-margin softmax \cite{liu2016large} & 2016        & 512        & 98.71 &   -    \\ \hline
SphereFace \cite{liu2017sphereface} & 2017          & 512      & 99.42 & 95    \\ \hline
Neural Aggregation \cite{yang2016neural} & 2017        & 128      &  -   & 95.72 \\ \hline
\end{tabular}
\end{table}

\section{Standardization of Deep Feature Descriptor}

In the context of video big data, to further ensure interoperability in deep learning based video analysis, a standard that focuses primarily on defining the syntax of compressed deep feature descriptors is essential. This section clarifies the
issues to be solved in the standardization process and how they might be pragmatically
approached. We believe that such AI oriented standard could represent a sea change in the future smart city applications.

\subsection{Compact Deep Feature for Video Analysis}
As introduced in Section 3, the features extracted by deep neural networks are gradually replacing the handcrafted features in many visual intelligence analysis. Due to millions of parameters lying in the deep network, as well as a series of non-linear mappings, the deep network can present high discrimination capability with pretty lower memory costs compared with the handcrafted features. Moreover, when massive training data is available, the involvement of end-to-end learning scheme would further sharpen the feature discrimination ability. Here, we investigate the performance and feature compactness of the recent remarkable works in four typical analysis tasks in city surveillance.

\begin{figure*}[t]
\centering
\includegraphics[width=6.3in]{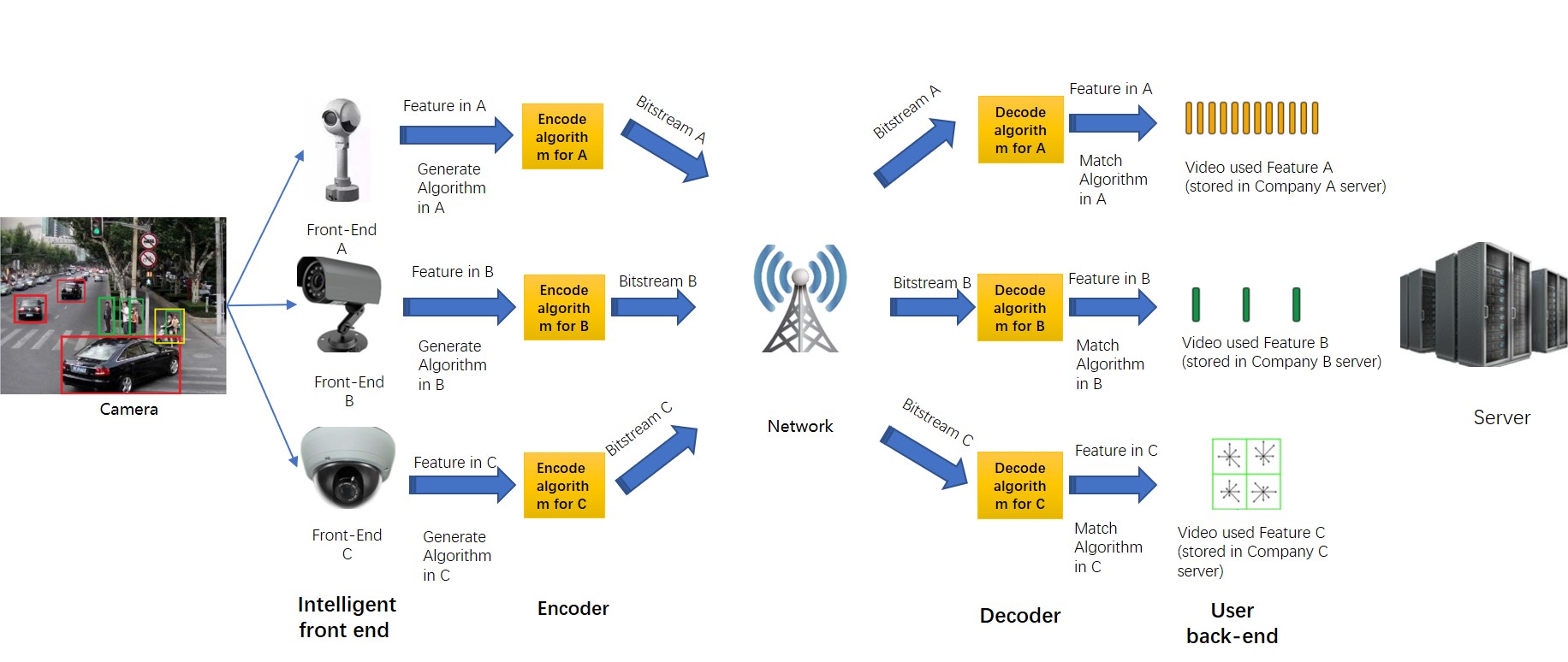}
\caption{Illustration of traditional feature transmission framework. Different feature bitstreams originated from different front end cameras have their own organization and syntax, such that the corresponding encoding and decoding algorithms should be performed at local and server ends, respectively.}
\label{fig:framework1}
\end{figure*}

\begin{figure*}[t]
\centering
\includegraphics[width=7.1in]{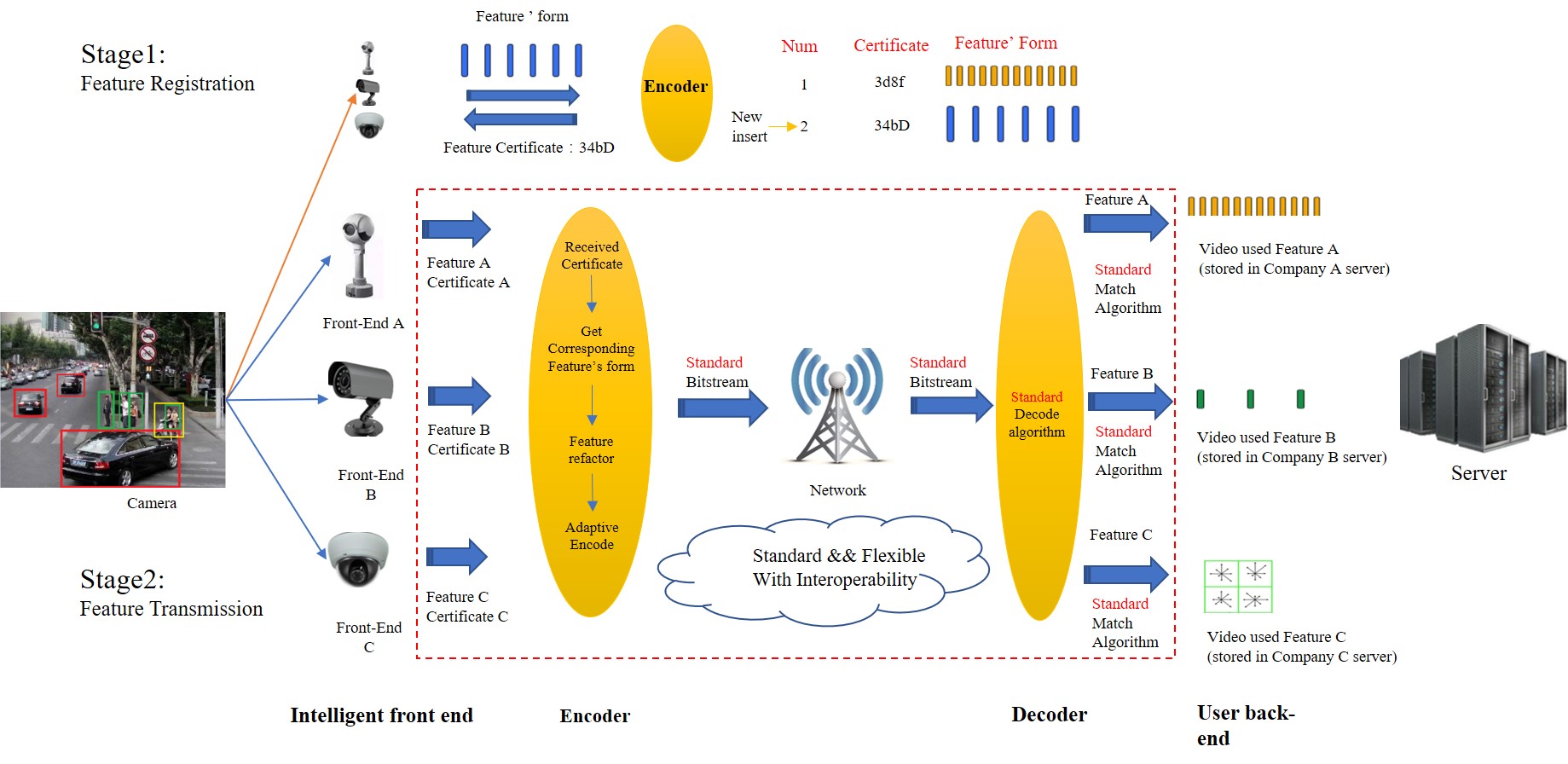}
\caption{Illustration of the interoperability-enabled feature transmission framework. First, new features are required to register its feature organization and syntax in the encoder and obtain the corresponding certificate. Second, features generated at the front end will be reorganized according to its certificate registered in encoder, and then encoded in standard bitstream syntax. Therefore, The server end can leverage standard algorithm to decode the received features }
\label{fig:framework2}
\end{figure*}

Although different network structures are employed in different analysis tasks, we find that the features can be uniformly represented without significantly sacrificing the analysis accuracy.
Table \ref{cdva} lists the video retrieval performance of the deep learning features with off-the-shelf CNN model reported in CDVA benchmarks. From the perspective of the performance and feature compactness, the deep learning feature shows the competitive performance. With the advance of network structure and training scheme, the performance of deep features have also been dramatically improved, being state-of-the-art on several image retrieval benchmarks, such as Holiday, Oxford5K and Paris. The person/vehicle ReID tasks also benefit a lot from the success of deep networks, as shown in Tables \ref{vehicleid}\&\ref{person}. In particular, we observe there is a trend that the recent methods employ features with much lower dimensions to produce the representation of an object, with the help of more powerful networks and well-defined loss functions. Similar trends can also be found in the recent efforts for face recognition, as shown in Table \ref{face}. In addition, the performance improvements originate from the introducing of object proposal in image retrieval has been witnessed in several benchmarks such as Oxford and Paris. The performance gains originate from the better localization ability provided by region proposal as shown in Fig. \ref{retrieval1} (c)\&(d). However, the obvious demerit of region-based methods is the relatively high feature dimension due to the representation requirements of multiple targets in query or reference images. As such, the region proposal mechanism would be practically applicable with the representations of compact feature. In Fig. \ref{retrieval1} (a)\&(b), the performance variations against the dimension are illustrated. All these experimental results demonstrate that these tasks can be successfully achieved with identical or similar feature dimension. For example, when the feature size reaches 512 dimensions, in most of the face recognition cases, competitive results can be obtained. Obviously, similar phenomenon can also be found when the dimension reaches 512 in CDVA, 512 in Person ReID, and 1024 in vehicle ReID.  In a word, a converging point can be feasibly attained from the perspectives of feature dimensions, proposal numbers, network structure. 
Another observation is that the performance variation along with the augment of proposals can also arrive at a saturation eventually as shown in Fig. \ref{retrieval1} (c)\&(d). Such observations provide useful evidence for the further standardization of deep learning features, as discussed in Section 4.2. 
In the future, it is also expected that more compact and discriminative feature representations will emerge due to the advance of the network architectures and optimization strategies.

\subsection{Toward Standardization of Deep Features}

It is apparent that the compact deep feature possesses many favorable properties for the applications of the smart city. However, the explosion of the deep learning models is also creating many challenging research problems. In particular, it is worth noting that the feature coding differs with traditional video coding in that an end-to-end feature coding pipeline involves both feature extraction and compression. In other words, for video coding the source visual signals are established and available, i.e. the pixel values. By contrast, in feature coding, different deep learning models would create dramatically different features for the subsequent compression process. Therefore, a complete and exhaustive standard that can fully ensure the interoperability typically specifies the standardization of both feature extraction and compression. As such, any bitstreams that conform to such standard can be meaningfully compared. 


Such standardization requires the deterministic deep network model and parameters. Nevertheless, the recent research achievements of deep learning emerge in endlessly, and moreover, there is a lack of the generic deep model that can be applied to a broad of tasks in video surveillance. Therefore, the standardization of deep learning model is not ready for prime time.
Here, we propose the concept of semi-interoperability for feature coding, which only standardizes the feature compression. In other words, only the pipeline from raw features to the compressed bitstream is taken into consideration, and the final syntax that specifies the compact deep features is standardized. The raw feature extraction process is left open for future exploration. Such strategy is based on the key observation that the raw features for these tasks can be uniformly represented, as demonstrated in Section 4.1. 
As such, the increasing demand for the interoperability in smart city and the explosion of deep learning techniques can be well balanced.

The semi-interoperability based standardization strategy is dual to the video coding standard where only the decoder is standardized. The decoder conforming to the standard can only correctly recover the features, but does not account for the explanation of the features as the deep learning model is not specified. Therefore, such strategy only ensures that any deep learning feature bitstreams from the same deep learning model conforming to such standard can be meaningfully matched after decoding. In other words, it does not fully support the interoperability and bitstreams conforming to such standard may convey different information. On the other hand, the advantage lies in that in the future any effective deep learning models can seamlessly collaborate with this standard, such that the standard can be kept with long-lasting vitality. Moreover, though there are multiple tasks in video surveillance and each task corresponds to the specific deep learning model, as long as the final generalized bitstreams from these models conform to the standard, they can be successfully decoded by a unified decoder. Here, the traditional feature compression and standardized feature compression frameworks are shown in Figs.~4\&5. It is observed that the bit-streams from different ends can be uniformly represented and transmitted, such that a unified decoder can be used to decode such bitstream to enable the semi-interoperability. 

Regarding feature compression, the high redundancy of deep learning features in video sequences needs to be removed. In particular, many video coding technologies can be analogously transferred to feature codings, such as inter prediction, intra prediction and rate distortion optimization. In addition, since the basic role of video surveillance is to analyze and explain the object behaviors, and in many occasions within a video frame there are multiple objects, it is natural to extend the frame level feature extraction and compression to the object level based on object proposals. For example, real-time object detectors such as YOLO \cite{redmon2016you} can be adopted to localize the target objects such as persons, vehicles, heads or other objects of interest, then the regions of interest will be feed into the corresponding networks designed for specific tasks to obtain the feature representation.
This also requires the non-local intra prediction to remove the redundancy from different objects within a frame. As such, how these redundancies can be removed and how the final bitstream is composed of should be further investigated in the standardization exploration. 

It is also anticipated that in the future the deep learning models are developed to maturation and generic as well as dynamic feature representations can be learned from surveillance videos. At that stage, there may emerge a unified deep learning model that can be standardized to achieve the full interoperability. Generally speaking, such deep learning model can not only deal with the various video surveillance tasks, but also enjoy the properties such as lightweight and friendly for implementation. Overall, the message we
are trying to send here is not that the standardization of deep model for feature extraction is abandoned. 
Rather, we hope to make the point that at the current stage, there are flexible and practical alternative solutions for the standardization
that can be deployed. 

\section{Outlook}

We have discussed the practical issues and envisioned the future standardization of deep learning features in the context of large-scale video management in the smart city.
Rather than exhaustively establishing the whole feature representation process including both extraction and compression, we have emphasized on the great potentials of standardizing the bitstream syntax of the compressed features. Such strategy is significantly different from the previous MPEG-7 visual standards such as CDVS and CDVA, and the deep learning models are not required to be specified to conform to the standard, which further enhances the flexibilities in the proliferation of deep learning technologies. In the future, it is expected that such AI oriented feature coding standard can play important roles in the establishment of the visual system of the city brain, and impact the new development of future AI technologies.

\bibliographystyle{IEEEbib}
\bibliography{refs1,refs}

\end{document}